\documentclass[sigconf]{acmart}
\usepackage{enumitem}
\usepackage{multirow}
\usepackage{multicol}
\usepackage{tabu}
\usepackage{graphicx}
\usepackage{caption}
\usepackage{subcaption}
\usepackage{bm}
\AtBeginDocument{
  \providecommand\BibTeX{{
    \normalfont B\kern-0.5em{\scshape i\kern-0.25em b}\kern-0.8em\TeX}}}

\setcopyright{acmcopyright}
\copyrightyear{2020}
\acmYear{2020}
\acmDOI{TBA}

\acmConference[LegalAI’20@SIGIR’20]{International ACM SIGIR Conference on Research and Development in Information Retrieval}{July 30, 2020}{Virtual Event, China}
\acmBooktitle{43rd International ACM SIGIR Conference on Research and Development in Information Retrieval, LegalAI’20@SIGIR’20, July 30, 2020, Virtual Event, China}
\acmPrice{15.00}
\acmISBN{TBA}

\begin{document}

\title{AI-lead Court Debate Case Investigation}

\author{Changzhen Ji}
\affiliation{
  \institution{Harbin Institute of Technology}
  \city{Harbin}
  \country{Heilongjiang, China}
}
\email{czji_hit@outlook.com}

\author{Xin Zhou}
\affiliation{
  \institution{Alibaba Group}
  \city{Hangzhou}
  \country{Zhejiang, China}
}
\email{eric.zx@alibaba-inc.com}

\author{Conghui Zhu}
\affiliation{
  \institution{Harbin Institute of Technology}
  \city{Harbin}
  \country{Heilongjiang, China}
}
\email{conghui@hit.edu.cn}

\author{Tiejun Zhao}
\affiliation{
  \institution{Harbin Institute of Technology}
  \city{Harbin}
  \country{Heilongjiang, China}
}
\email{tjzhao@hit.edu.cn}










\begin{abstract}
The multi-role judicial debate composed of the plaintiff, defendant, and judge is an important part of the judicial trial. Different from other types of dialogue, questions are raised by the judge, The plaintiff, plaintiff's agent defendant, and defendant's agent would be to debating so that the trial can proceed in an orderly manner.
Question generation is an important task in Natural Language Generation. 
In the judicial trial, it can help the judge raise efficient questions so that the judge has a clearer understanding of the case.
In this work, we propose an innovative end-to-end question generation model-Trial Brain Model (\textbf{TBM}) to build a Trial Brain, it can generate the questions the judge wants to ask through the historical dialogue between the plaintiff and the defendant. 
Unlike prior efforts in natural language generation, our model can learn the judge's questioning intention through predefined knowledge.
We do experiments on real-world datasets, the experimental results show that our model can provide a more accurate question in the multi-role court debate scene.

\end{abstract}



\maketitle

\section{Introduction}

The contradiction between the gradual increase of people's demands in pursuing social justice and relatively scarce public resources is one of the prominent contradictions in the current society. In a legal context, a lengthy and expertise-demanding trial can be a high threshold for a litigant, while the judge has to spend significant efforts to investigate the case and explore exhaustive questionable factors. This can be very challenging for junor judges, while a careless negligence can bring unforgivable consequences. Unfortunately, federal/district court judges are experiencing daunting workload, e.g., statistics show that the typical active federal district court judge closed around $250$ cases in a year \cite{trac_report_1,trac_report_2}. Applying novel artificial legal intelligence techniques to facilitate the lawsuit process so as to alleviate the information overload for judges is of great significance.

In a longer term, this pioneer investigation can provide critical potentials to enable automatic trail processing system so as to improve trial efficiency. For junior judges, such automation accompanying can provide important assistance during an ongoing trial debate or court rehearsal.  

\begin{table}
  \caption{Example Dialog in Court Debate Dataset}
  \label{tab:table11}
 \begin{tabular}{l|p{6cm}}
     \toprule
      \textbf{Role} & \textbf{Dialogue} \\
    \midrule
      ...        &                 ...\\
      Judge  &  Defendant, is there any evidence to provide to the court?\\
      Defendant   &   No.\\
      Judge  &  Plaintiff, what's your relationship with <person>?\\
      Plaintiff   &   Friends.\\
      Judge  &  What did <person> borrow money for at that time?\\
      Plaintiff   &   To operate a supermarket.\\
      Defendant   &   I heard the people say they were in the same circle.\\
      Judge  &  Which circle?\\
      Defendant   &   Gambling circle.\\
      Judge  &  Plaintiff, do you know whether <person> participated in gambling?\\
      Plaintiff   &   I don't know. I'm not with him.\\
      Judge  &  When <person> borrowed money from you, did you agree that it was his personal debt?\\
      Plaintiff   &   No.\\
      Judge  &  Defendant, who paid your living expenses with <person>?\\
      Defendant   &   It is my expenditure.\\
      Judge  &  Who pays for the family expenses?\\
      Defendant   &   Me.\\
      ...  &   ...\\
    \bottomrule    
\end{tabular}
\end{table}

\begin{figure*}
    \centering
\includegraphics[height=0.38\textheight]{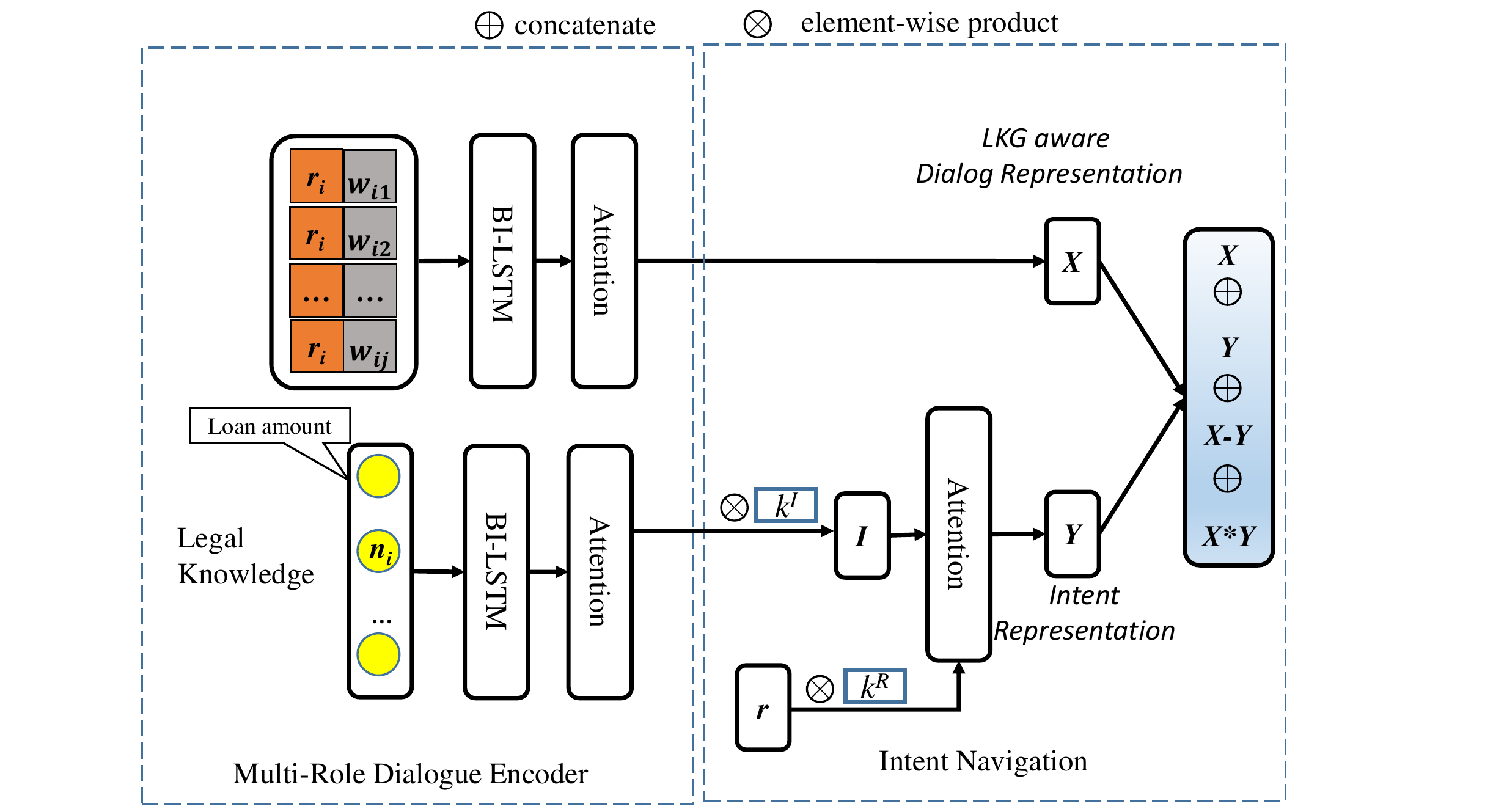}
    \caption{Network Architecture of the Proposed Method}
    \Description{figure1Description}
    \label{fig:model-overview}
\end{figure*}

Unlike existing one-to-one dialogue systems, court debate is conducted under multi-role scenario, and an experienced judge may propose the next question with respect to the following factors. First, the prior debate context can be important to decide the content of the forthcoming question (algorithmically, debate context characterization can be essential for this task). Second, response's role information is of significant, i.e., same content from different parties can deliver very different information (algorithmically, role information should be integrated into the debate context representation learning). Third, judge's intention can scope the content of the forthcoming question. For instance, the judge may raise several related question (e.g., \texttt{when did you get divorced?}, \texttt{who raise the child?}, \texttt{who cover the daily household expense?}) in te- rms of the current intent (e.g., to confirm if the spouse should be also responsible for the debt). After the judge collected enough information of the current intent, he/she can switch to the next intent (e.g., to check if there is actual agreement on the term of the loan).

we propose an innovative end-to-end question generation model to build a trial brain, it can generate the questions the judge wants to ask through the historical dialogue between the plaintiff and the defendant. Unlike prior efforts natural language generation, our model can learn the judge's questioning intention through predefined knowledge. Fig. \ref{fig:model-overview} depicts the systematic structure of the proposed model. 

To sum up, our contributions are as follows:
\begin{enumerate}

\item The proposed model is able to learn the judge intention transition for question generation navigation through predefined knowledge.
\item The proposed model can provide a more accurate question in the multi-role court debate scene.

\end{enumerate}

\section{Problem formulation}
Let $D$ denote an arbitrary dialogue fragment, containing $L$ utterances. Each utterance $U_i$ in $D$ is composed of a sequence of $l$ words (namely sentence) $S_i$ along with the associated role (of the speaker) $r_i$. We define the last question in $D$ raised by the judge as $U^q$ and its historical conversations is denoted as $D^-=\{U_1,U_2,...,U_n\}$. In the task of question generation, the proposed algorithm can generate $U^q$ given a corresponding $D^-$.

To be clarified, the definition of important notations in the following sections are illustrated as follows:
\begin{itemize}[leftmargin=2em]
\setlength{\itemsep}{0pt}
\setlength{\parsep}{0pt}
\setlength{\parskip}{0pt}
    \item $D$: a debate dialogue fragment containing $L$ utterances;
    \item $r_i$: the role of the speaker in $U_i$ (i.e. judge, plaintiff, defendant and witness);
    \item $S_i$: the text content of $U_i$;
    \item $S^q$: the text content of $U^q$;
    \item $D^-$: the historical conversations of $U^q$;
    \item $I_i$: the intent of utterance $U_i$;
\end{itemize}

Note that $\mathbf{U_i}$, $\mathbf{r_i}$, $\mathbf{S_i}$, and $\mathbf{I_i}$ represent the embedding representations of the corresponding variables in the list.

\section{Model}

The innovative multi-view utterance encoder characterizes three kinds of information -  role embedding, semantic embedding, and legal knowledge embedding. More importantly, intention, as an important latent variable, navigates the direction of question generation, which is coached by LKG knowledge transformation. Finally, we use pointer generator networks \cite{see2017get} further enhance the quality of question generation.

\subsection{Multi-Role Dialogue Encoding}
\label{Sec:dialogue_encoder}

\subsubsection{Role Representation}
\label{Sec:role_representation}
In term of role embedding, we use dense vectors to represent different roles  (e.g., presiding judge, plaintiff, defendant and witness) in the debate dialogue.

\subsubsection{Semantic Representation}
\paragraph{\makebox[1em][l]{$\,\bullet$}Utterance Layer}
In the utterance layer, we utilize a Bidirectional Long-Short Term Memory networks (Bi-LSTM) \cite{hochreiter1997long} to encode the semantics of the utterance while maintaining its syntactics. 

\paragraph{\makebox[1em][l]{$\,\bullet$}Dialogue Layer}
To represent the global context in a dialogue, we use another Bi-LSTM to encode the dependency between utterances to obtain a global representation of an utterance as dialogue representation, denoted as $X$.




\subsubsection{Legal Knowledge Representation}
Legal knowledge is an element marked by the judge, such as (borrowing time, loan amount.....).  We also use dense vectors to represent different element, and then encode it by LSTM, expressed as $\mathbf{\overline{h^p}}$.

\subsection{Representation Enhancement via Intent Navigation}

\label{Sec:Intent_navigation}
When judge construct the forthcoming question, he/she should consider three kinds of information: (1) the intent of the question, (2) the content of the question, and (3) the litigant role being asked. It motivates us to learn the intent transition to navigate the forthcoming question generation. To represent the intent of judge, we rely on the Legal Knowledge to learn the navigation among different sequence of legal concepts (see Eq. \ref{Eq:intent}). At the same time, we learn a role transfer matrix to represent the role to be asked, in other words, the role who answer the generated question (see Eq. \ref{Eq:role}). Note that the role used in Sec. \ref{Sec:role_representation} is speaker's role of the corresponding utterance while the role mentioned here is the responser's role of the current utterance\footnote{We mainly focus on the response-role of judge's question since commonly the next role of litigant's answer will be always the judge.}



\begin{equation}
\label{Eq:intent}
\mathbf{I} = \sigma(k^I*\mathbf{\overline{h^p}})
\end{equation}

\begin{equation}
\label{Eq:role}
\mathbf{R} = \sigma(k^R*\mathbf{r})
\end{equation}

The two parameters $k^I$ and $k^R$ stands for the learnable hidden matrix for simulating the intent transfer and the response-role transfer respectively, in which the matrix elements are all values between 0 and 1. 




We merge intention information and next role information as below:
\begin{equation}
\mathbf{H} = ([\mathbf{I_1},\mathbf{R_2}], [\mathbf{I_2},\mathbf{R_3}],...,[\mathbf{I_i}, \mathbf{R_{i+1}}])
\end{equation}

where $\mathbf{R_{i+1}}$ represents the next role of the current utterance. 

We further compress the original redundant information and assign more weight to important information via attention mechanism:
\begin{equation}
\mathbf{Y} = {\sum_{i=1}^n}{\frac {exp(\mathbf{I_i})} {\sum_{i=1}^n exp(\mathbf{I_i})}}*\mathbf{H}
\end{equation}

Next, we fuse the original information with the intent/role transformation information:
\begin{equation}
\mathbf{Z} = [\mathbf{X}, \mathbf{Y}, \mathbf{X}*\mathbf{Y}, \mathbf{X}-\mathbf{Y}]
\end{equation}

\subsection{Parameter Optimization}
In question generation learning, for each dialog $D$, we use cross-entropy to formulate the problem as follows:

\begin{align*}
& loss = -\log P(S^q\mid D) \\
& \quad \  \, =-\sum_{j=1}^{l}\log P(w_{ij}\mid w_{i1:j-1},D)
\end{align*}

Denoting all the parameters in our model as $\delta$. 
Therefore, we obtain the following optimized objective function:

\begin{equation}
\underset{\theta}{min}\; loss = loss + \lambda\left \| \delta \right \|_2^2
\end{equation}

To minimize the objective function, we use the diagonal variant of Adam in \cite{zeiler2012adadelta}. At time step $t$, the parameter $\delta$ is updated as follows:

\begin{equation}
\delta_t \leftarrow \delta_{t-1} - \frac{\mu}{\sqrt{\sum_{i=1}^tf_i^2}}f_t
\end{equation}
where $\mu$ is the initial learning rate and $f_t$ is the sub-gradient at time $t$.

According to the evaluation results on the development set, all the hyperparameters are optimized on the training set.



\section{Experimental Settings}
\subsection{Dataset}
\label{Sec:dataset}
In the experiment, we collected $136,019$ court debate records of civil Private Loan Disputes cases, from which we randomly extracted\footnote{We only selected the fragments in which there are at least five historical utterances as context for our task of next question generation.} $302,650$ continuous dialogue fragments as independent samples for training, developing and testing\footnote{The entire dataset is divided by a ratio of 8:1:1 for training, developing and testing, respectively.}. In total, it contains more than $4$ million sentences and each dialogue fragment, on average, contains $13.38$ sentences. The details of the dataset is illustrated in Table \ref{tab:table1}.


\begin{table}[!htbp]
  \caption{Statistics of the Processed Dialogue Fragments}
  \label{tab:table1}
  \begin{tabular}{cccc}
    \toprule
    Dataset & \#Samples & \#Utterances & \#avg\_length\\
    \midrule
    train & 242,120 & 3,238,956 & 13.38\\
    test & 30,265 & 404,833 & 13.38\\
    development & 30,265 & 404,870 &13.38\\ 
    \textbf{Total} & \textbf{302,650} & \textbf{4,048,659} & \textbf{13.38} \\
  \bottomrule
\end{tabular}
\end{table}

\subsection{Baselines}
In order to demonstrate the validity of our model, we selected some traditional classical methods and the latest mainstream methods for text generation. 
The tested baselines are illustrated as follows:
\begin{itemize}[leftmargin=2em]
\item \textbf{LSTM}\cite{hochreiter1997long}: We replace all bidirectional LSTM with LSTM in our proposed model.
\item \textbf{ConvS2S} \cite{gehring2017convolutional}: LSTM be replaced CNN in the encoder. 
\item \textbf{ByteNet} \cite{kalchbrenner2016neural}: It is a one-dimensional convolutional neural network that is composed of two parts, one to encode the source sequence and the other to decode the target sequence.
\item \textbf{S2S+attention} \cite{nallapati2016abstractive}: The Seq2Seq framework relies on the encoder-decoder paradigm. The encoder encodes the input sequence, while the decoder produces the target sequence. Attention mechanism is added to force the model to learn to focus on specific parts of the input sequence when decoding.
\item \textbf{PGN} \cite{see2017get}: It is another commonly used framework for text generation which enables copy mechanism to aid accurate reproduction of information, while retaining the ability to produce novel words through the generator.
\item \textbf{Transformer} \cite{vaswani2017attention}: A neural network architecture based on self-attention mechanism.
\end{itemize}

\subsection{Evaluation Metrics}
To automatically assess the quality of the \textit{generated question}, we used ROUGE \cite{lin2003automatic} and BLEU \cite{papineni2002bleu} scores to compare different models. We report ROUGE-1, ROUGE-2, ROUGE-3 as the means of assessing informativeness and ROUGE-L as well as BLEU-4 for assessing fluency. 


\section{Result discussion}
\begin{table}[!t]
  \caption{Main Results of All Test Methods. Note that the results show in \textbf{TBM(our)} rows are statistically significant different from the corresponding value of all the baseline models ($p$-value$<0.001$).}
  \label{tab:table2}
  \begin{tabular}{c|ccccc}
    \toprule
    \textbf{Model} & \textbf{R.-1} & \textbf{R.-2} & \textbf{R.-3} & \textbf{R.-L} & \textbf{BLEU}\\
    \midrule
    LSTM           & 29.33  &  14.97  &  10.34  &  26.65  &  11.59 \\
    ByteNet        & 35.57  &  19.47  &  14.27  &  32.56  &  17.73 \\
    ConvS2S        & 36.35  &  20.42  &  15.98  &  33.03  &  17.97 \\
    S2S+attention  & 36.54  &  20.72  &  16.40  &  33.29  &  17.96 \\
    PGN            & 37.67  &  21.93  &  17.42  &  34.39  &  18.75 \\
    Transformer    & 37.59  &  23.26  &  18.71  &  35.38  &  18.58 \\
    \hline
    \textbf{TBM(our)} & \textbf{39.02}  &  \textbf{24.56}  &  \textbf{21.03}  & 	\textbf{38.12}  &  \textbf{24.17} \\
  \bottomrule
\end{tabular}
\end{table}


\begin{figure}[!t]
\centering
  \centering
  \includegraphics[width=\linewidth]{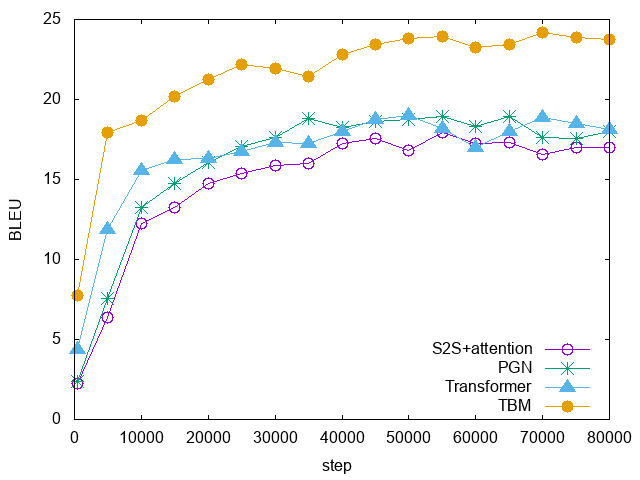}
  \label{fig:converge-1}

\caption{The performance of tested methods.}
\vspace{-8pt}
\label{fig:convergence}
\end{figure}

The performance of all tested methods is reported in Table \ref{tab:table2}. As the upper part of Table \ref{tab:table2} shows, the proposed method \textbf{TBM} is significantly ($p$-value$<0.001$) superior than all the tested baselines over all the evaluation metrics. 

\section{Conclusion and outlook}
Dialogue generation has been well studied in NLP.
At present, some achievements have been made. 
However, it is also faced with great challenges.
Many superior models only achieve good results in specific field.
In this paper, we define a new task of the question generation in judicial trial. The proposed Trial Brain Model can learn the judge's questioning intention through predefined knowledge. Judge-centered debate context heterogeneity is the landmark of this model, i.e., a delicately designed multi-role dialogue encoding mechanism via Legal Knowledge with the representation enhancement through intent navigation by simulating the intention switch across different conversations. The empirical findings validate the hypothesis of this bionic design for judge logic reduction. An extensive set of experiments with a large civil trial dataset shows that the proposed model can generate more accurate and readable questions against several alternatives in the multi-role court debate scene.

\section*{Acknowledgments}
This work is supported by National Key R\&D Program of China 
(2018YFC0830200;2018YFC0830206).


\bibliographystyle{www}
\end{document}